\documentclass[12pt, titlepage]{article}
\usepackage{ae}
\usepackage{graphicx}
\usepackage{ulem}
\usepackage{hhline}
\usepackage{amsfonts}
\begin{document}
\bibliographystyle{alpha}
\title{Recorded Step Directional Mutation for Faster Convergence}
\author{Ted Dunning\\
Chief Scientist\\
Aptex Software}
\date{25 September, 1995}
\begin{abstract}

Two meta-evolutionary optimization strategies described in this paper
accelerate the convergence of evolutionary programming algorithms
while still retaining much of their ability to deal with multi-modal
problems.  The strategies, called directional mutation and recorded
step in this paper, can operate independently but together they
greatly enhance the ability of evolutionary programming algorithms to
deal with fitness landscapes characterized by long narrow valleys.
The directional mutation aspect of this combined method uses
correlated meta-mutation but does not introduce a full covariance
matrix.  These new methods are thus much more economical in terms of
storage for problems with high dimensionality.  Additionally,
directional mutation is rotationally invariant which is a substantial
advantage over self-adaptive methods which use a single variance per
coordinate for problems where the natural orientation of the problem
is not oriented along the axes.

Step-recording is a subtle variation on conventional meta-mutational
algorithms which allows desirable meta-mutations to be introduced
quickly.  Directional mutation, on the other hand, has analogies with
conjugate gradient techniques in deterministic optimization
algorithms.  Together, these methods substantially improve performance
on certain classes of problems, without incurring much in the way of
cost on problems where they do not provide much benefit.  Somewhat
surprisingly their effect when applied separately is not consistent.

This paper examines the performance of these new methods on several
standard problems taken from the literature.  These new methods are
directly compared to more conventional evolutionary algorithms.  A new
test problem is also introduced to highlight the difficulties inherent
with long narrow valleys.
\end{abstract}
\maketitle
\section{Overview}

\subsection{Arguments for Meta-evolution}

A number of stochastic optimization procedures have been developed
since the electronic computer has made automated optimization
possible.  Methods which have had substantial recent development
include simulated annealing, genetic algorithms, evolutionary
strategies and evolutionary programming.  One shared property of each
of these classes of algorithms is that they trade some degree of
convergence speed for a decreased likelihood of avoiding locally
optimal but globally suboptimal solutions.

In each of these well-known algorithms, there is a parameter or set of
parameters which can be manipulated to affect this trade-off between
convergence power and speed.  In simulated annealing, this parameter
is the simulated temperature, while in evolutionary programming, this
parameter is the mutation rate.  Typically, the temperature or
mutation rate is decreased as the optimization progresses.  This
tactic substantially improves the rate of convergence, often without
significantly increasing the likelihood of finding a suboptimal
solution.  In special cases such as a quadratic bowl, cooling
schedules can be derived which satisfy various theoretical constraints
regarding the effort needed to have a given probability of finding a
solution in a given amount of time, but this cannot be done in general
since the derivation of the optimal annealing schedule requires
detailed knowledge of properties of the function being optimized.
Instead, an arbitrary and hopefully sufficiently conservative cooling
schedule is typically invented and used.

An alternative to a fixed cooling schedule is to derive the cooling
schedule adaptively as the optimization algorithm learns about the
fitness landscape that it is exploring.  The idea that the mutation
rate itself should be a parameter specific to each member of the
population to be evolved is not new and has been recently explored in
\cite{fogel92} and \cite{fogel}.  This form of meta-evolution is
attractive in that no explicit cooling schedule need be given.

\subsection{Common problems}

Problems whose solutions are found in long narrow valleys cause severe
problems with evolutionary programming algorithm because the
narrowness of the valley greatly decreases the probability of finding
a solution which improves on a point which is already on the floor of
the valley.  These problems have been attacked in the past by using a
covariance matrix to cause mutations to be correlated as described in
\cite{sebald}, \cite{schwefel} and \cite{fogel92}.  This method is
similar in essence to the conjugate gradient techniques used in
conventional numerical optimization codes in that they concentrate the
exploration of the fitness landscape in particular directions based on
past experience.  Various forms of directional mutation has been the
subject since the mid 60's as indicated by the work of Bremermann and
others \cite{bremermann64,bremermann65}.

Combining this method with meta-evolution as is done in the methods
presented in this paper raises some interesting problems, however.  In
particular, in meta-evolution, all parameters which control mutation
are included in the genome and must themselves be subject to mutation.
But, if there are $n$ real valued parameters in the genome initially,
then it takes $n^2$ entries in a covariance matrix to describe how
those $n$ parameters should be changed.  Including the covariance
matrix in the genome increases its size to $n^2+n$ real valued
parameters and raises the serious question of how the mutation of the
new parameters should be described.  Self-similar mutation schemes can
avoid the risk of a recursive explosion in the number of parameters to
be optimized.

Even so, the original $n^2$ covariance parameters in the description
of each element of the entire population can be very expensive, even
if meta-meta-mutation parameters are not included.  Thus, it is
desirable to find a more economical method for describing correlated
mutation, and to find a way so that this correlated mutation
description contains its own description of how it should be changed.
One additional desiderata is that the meta-mutation be self similar so
that the algorithm is insensitive to changes in scale.  The two new
strategies described in this report address this need.  Other work
along these lines can be found in \cite{fogel97} where a directional
mutation scheme is described which has overhead similar to the methods
described here.

It should be noted that the stochastic optimization methods which have
used Cauchy distributed mutations as in
\cite{fishman,szu,yao91,yao95,yao97} are inherently not rotationally
invariant.  In evolutionary optimizations dealing with a large number
of dependent parameters, this means that the proportion of progeny
which change only a few parameters will be much higher than would be
expected under the assumption of complete rotational symmetry of the
optimization algorithms.  For many problems, notably including the
synthetic problems often used to evaluate optimization algorithms,
this coordinate orientation can be highly advantageous.  For example,
in the multi-modal test problems explored in \cite{yao97} the local
optima are arranged in orthogonal grids parallel to the parameter
axes.  In other problems of significant practical significance,
however, correlated changes in parameters are necessary.  Examples of
this need for correlated changes occur in artificial neural networks
or in electronic filter design.

\section{Two New Strategies}

\subsection{Step Recording}

In a conventional meta-mutation algorithm, the mutation rate for each
member of the population is mutated independently of the state vector.
This method can lead to slower convergence, especially in conjunction
with directional mutation if a low probability step leads to
improvement in fitness.  If this happens, repeating a step like the
one that caused the improvement can be advantageous.  If the mutation
rate (and possibly the mutation direction) was changed independently
of the fitness parameters, then similar steps will probably stay
unlikely and convergence will be slow.

This situation will happen when a very fit solution in a small basin
has been found due to taking a very large step.  That particular
solution will tend to remain in the population, but further
improvement is unlikely unless subsequent small steps are taken.  Even
if an improved solution is found by taking a small step (which is
unlikely, but it will happen eventually), it is likely that the
mutation rate will still be large (which is why we found this solution
in the first place), and further improvements will only come slowly as
solutions with {\em both} better fitness and lower mutation rates are
found.  The typical sequence is for the mutation rate to decrease
first which then allows smaller steps to be taken resulting in further
optimization.

With step recording, on the other hand, the mutation rate of an
offspring is set to the magnitude of the distance between the parent
and child solution.  This coupling of mutation rate and change in
position means that any solutions which improve fitness by taking
small steps will automatically lead to a line of progeny which tend to
explore by taking small steps.  Similarly, when directional mutation
is combined with step recording, once a step is taken along the
fitness gradient, further steps similar to that one are likely.  This
provides algorithms using directional mutation a sense of history in
much the same manner as conjugate gradient methods use past history to
improve further optimization efforts.

\subsection{Directional Mutation}

In order to fully characterize all of the possible correlations
between $n$ random variables, it is necessary to use roughly $n^2$
quantities.  It does not, however, follow that a description this
complete is necessary to gain the benefits of directional mutation.
In particular, a covariance matrix specifies much more than just
mutation biased in a particular direction.  Indeed, a covariance
matrix contains much more information than can be reliably extracted
from the recent pedigree of a single member of a population.

Instead, we propose the use of a much more limited model of
directional mutation in which the mutation rate has a directional
component and an omni-directional component.  The total mutation is
the sum of mutations derived from each of these components.  The
directional component of mutation is restricted to a line, while the
omni-directional component is sampled from a symmetric gaussian
distribution.  Together these components give a total mutation
distribution which is an ellipsoidal gaussian distribution.  In a
heuristic attempt to enhance convergence, the directional component is
also biased slightly.

If we use the notation $N(\mu, \sigma)$ to indicate a normally
distributed random variable with mean $\mu$ and standard deviation
$\sigma$ and use $U(a, b)$ to indicate a uniformly distributed random
variable taken from the half open interval $[a, b)$, then the
following mutation algorithm suffices to provide a directional
mutation of the vector $\bf x$

\begin{displaymath}
\lambda := N(1,1)
\end{displaymath}
For each $x_i$,
\begin{displaymath}
x_i := x_i + N(0, \sigma) + \lambda k_i 
\end{displaymath}

Here $\lambda$ is a biased random variable which indicates how far to
go along the direction indicated by $\bf k$, while $\sigma$ provides
the magnitude of the omni-directional mutation component.

The mutation parameters $\bf k$ and $\sigma$ can themselves be mutated
by setting

\begin{displaymath}
\sigma := - (\sigma + |{\bf k}|/10) \log (1-U(0,1))
\end{displaymath}
\begin{displaymath}
\lambda := N(1,1)
\end{displaymath}
and then for each $k_i$,
\begin{displaymath}
k_i := N(0, \sigma) + \lambda k_i 
\end{displaymath}

In this algorithm the mutation of $\sigma$ is done by taking a new
value from the exponential distribution with mean equal to $\sigma$
augmented by a fraction of the magnitude of $\bf k$.  This cross
coupling between $\sigma$ and $\bf k$ prevents the mutations from
becoming too directional.  The mutation of $\bf k$ uses $\sigma$ to
provide diversity in direction and $\lambda$ to provide diversity in
magnitude in a manner identical with the way that the mutation of
$x_i$ was done.  The use of an exponential distribution is somewhat in
contrast with the trend in the literature toward the use of a
log-normal distribution, but the motivation is essentially the same.
Both the exponential and log-normal distributions allow self-similar
mutation which allows the entire algorithm to be scale invariant.

It should be noted that the meta-mutation operation described here is
self-similar and orientation independent.  This means that the
distribution of mutation parameters after several generations in the
absence of selection is invariant up to the scale and orientation of
the original value.  This fact also implies that the properties of the
resulting meta-evolutionary algorithm are subject to analysis by
renormalization methods.

To convert this algorithm to a step recording algorithm, the mutation
of $\bf x$ is simplified and is done after the mutation of $\sigma$
and $\bf k$ as shown below.

\begin{displaymath}
\sigma := - (\sigma + |{\bf k}|/10) \log (1-U(0,1))
\end{displaymath}
\begin{displaymath}
\lambda := N(1,1)
\end{displaymath}
\begin{displaymath}
k_i := N(0, \sigma) + \lambda k_i 
\end{displaymath}
\begin{displaymath}
x_i := x_i + k_i 
\end{displaymath}

The result is that $\bf k$ records the mutation step which was taken
so that if this step results in an improvement, similar steps are
likely to be used again.

Another notable feature of the algorithms described here are the
coupling between the directional parameters ${\bf k}$ and the
omni-directional parameter $\sigma$.  The coupling from ${\bf k}$ to
$\sigma$ allows a population to stop mutating directionally when
necessary as well as providing a bias which tends to increase the
overall mutation rate in the absence of selection for lower rates.
The coupling from $\sigma$ back to ${\bf k}$ allows changes in the
preferred direction of mutation to take place.

\section{Experimental Methods}

To provide a preliminary test of the efficacy of the proposed
algorithms, all four combinations of conventional meta-evolution,
meta-evolution with step recording, meta-evolution with directional
mutation and conventional meta-evolution with both step recording and
directional mutation were tested on three simple problems.  These
problems included a three dimensional symmetric quadratic bowl
(function F1 from \cite{fogel94}), a Bohachevsky multi-modal bowl
problem (function F6 from the same work) and a very narrow two
dimensional quadratic bowl whose axis was not aligned with either axis
(labelled F9 here to avoid conflict with F1 through F8 from
\cite{fogel94}).  These problems were not intended to provide a
comprehensive inventory of the interesting problems, but rather were
simply taken as exemplars which would highlight the contrast between
previous methods and the directional recorded-step method.  

The dimensionality of the test functions used here is quite low, but
the essential difficulty posed to previous evolutionary algorithms by
long narrow valleys which are not aligned along the coordinate axes is
independent of dimension.  Additional tests with dimensionality as
high as 30 show the same results as demonstrated here.

The test functions are described by the following functions:

\begin{displaymath}
f_1(x,y,z) = x^2 + y^2 + z^2
\end{displaymath}
\begin{displaymath}
f_6(x,y) = x^2 + 2 y^2 - 0.3 \cos(3 \pi x) - 0.4 \cos(4 \pi y) + 0.7
\end{displaymath}
\begin{displaymath}
f_9(x,y) = (x+y)^2 + (100 y - 100 x)^2
\end{displaymath}

For this test, the evolutionary algorithm used 20 survivors each
generation, each of which generated 9 progeny to create a population
of 200.  After evaluating the fitness function for each member of the
population, the entire population was sorted to find the best 20
members who would survive into the next generation.  

Each algorithm was run 10 times and a median fitness at each
generation was used to compare algorithms.  All programs were limited
to 50 generations or less.  Generally, convergence to a solution with
$10^{-8}$ of the correct value was found within far fewer generations.

\section{Results}

\begin{figure}
\centerline{\includegraphics{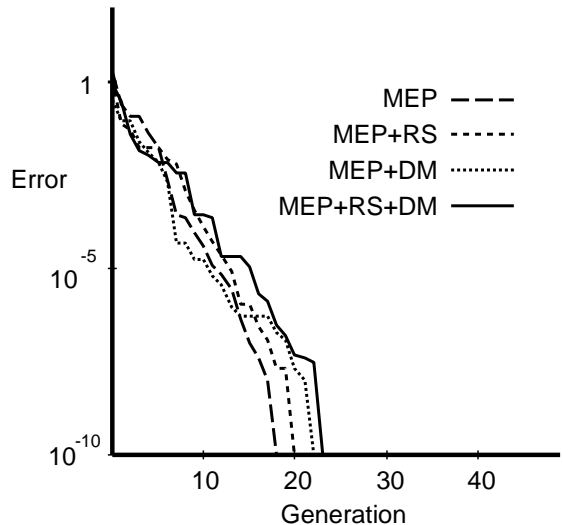}}
\caption{Convergence for Symmetric Bowl}\label{fig:f1}
\end{figure}

The graph in \ref{fig:f1} illustrates the convergence for the four algorithms
for the symmetric bowl (Function F1).
As can be seen, the convergence of the conventional meta-evolutionary
strategy is slightly faster than for the modified algorithms, but the
difference in terms of number generations required to converge is not
substantial and the ultimate accuracy of the final solution is
essentially identical.  It is interesting to note that the
omni-directional mutation rate was a close approximation of the square
root of the remaining error.  This behavior is close to the
theoretical optimum cooling for this problem; that it was derived
automatically by the meta-mutation was noteworthy.  Detailed
examination of the population showed that omni-directional mutation
was the dominant mechanism of exploration in the case of the symmetric
bowl.

\begin{figure}
\centerline{\includegraphics{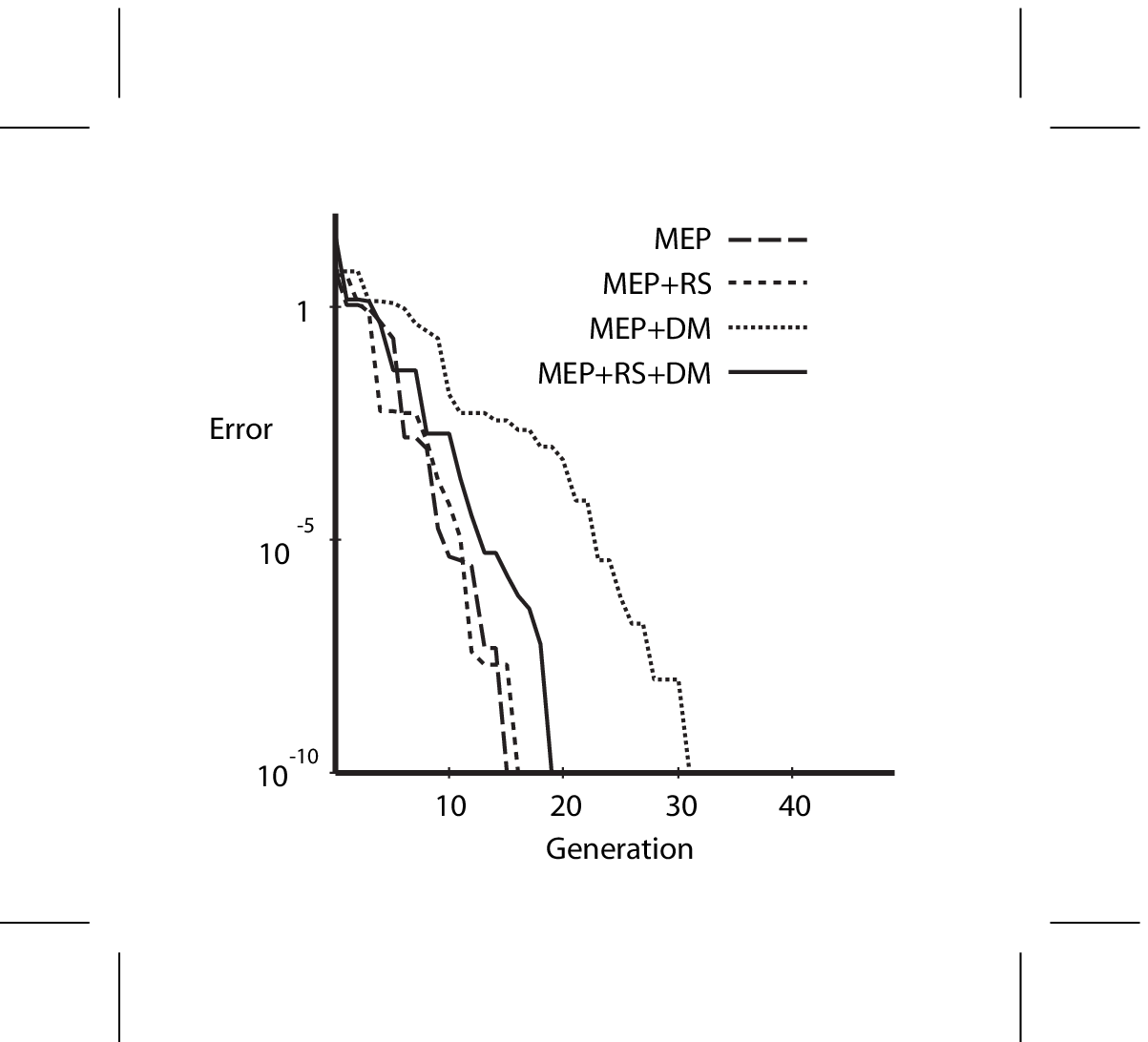}}
\caption{Convergence for Bohachevsky Function}\label{fig:f2}
\end{figure}

The graph in \ref{fig:f2} illustrates convergence for the Bohachevsky function.
Again, the difference between the algorithms is not striking, except
for the algorithm which used directional mutation without step
recording.  Even so, the degradation in convergence time was less than
a factor of two for directional mutation, and the loss in performance
for the other methods was minimal.

\begin{figure}
\centerline{\includegraphics{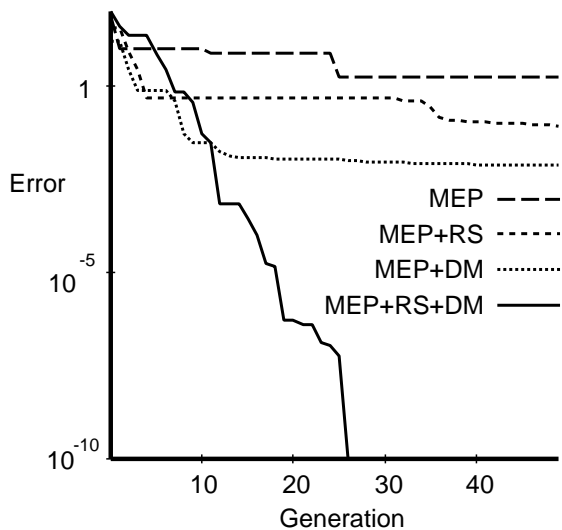}}
\caption{Convergence for Narrow Bowl}\label{fig:f3}
\end{figure}

Finally, the graph in \ref{fig:f3} illustrates the convergence rates for the
narrow quadratic bowl.

Here, all algorithms except for directional mutation with step
recording have severe problems with convergence.  The differences here
are highly significant.  The difference in the case of the
non-directional algorithms is due to the fact that with symmetric
mutation, if the mutation rate is much larger than the distance to the
major axis of the valley, then any mutation is likely to fall outside
of the narrow valley and thus not result in any improvement.  The
effect is that as the population approaches the major axis of the
valley, the mutation rate is decreased and progress toward the optimum
slows down.  Ultimately, solutions very close to the major axis are
found, and the mutation rate is reduced to a small value.  This low
mutation rate makes progress down the major axis of the valley toward
the global optimum quite slow.

It is not clear why directional mutation without step recording
performs so poorly, but early experiments with different meta-mutation
operators appeared to perform better, so the problem may have had more
to do with the meta-mutation itself than with an inherent defect in
the pure directional mutational algorithm.

The algorithm which used directional mutation and step recording
performed very well in the narrow valley problem.  Detailed
examination of evolving populations showed that populations far from
the major axis of the valley quickly evolved directional mutations
which took them to the major axis.  Once there, the populations
converted their directional mutations into omni-directional mutations
which were in turn converted into directional mutations strongly
oriented along the major axis of the valley.  This quickly lead to
bracketing of the solution at which point, the population variance
shrank rapidly.  Once the population had adapted to the nature of the
problem, convergence proceeded essentially identically to the
convergence behavior noted for the symmetric bowl problem ($f1$).

It is instructive to compare these results with those from the table
on page 173 of \cite{fogel94}.  The relevant parts of that table are
reproduced in \ref{tab:t1} and extended with the current results.
Note that the meta-evolutionary algorithms (the new columns which are
labelled MEP, MEP+RS and MEP+RS+DM) are clearly able to produce
results which are comparable with previous evolutionary algorithms
(the original columns which were labelled GA, DPE and EP in the
original work).  It should be remembered that when examining this
table that comparing the various forms of evolutionary programming
after such an extreme degree of convergence is not terribly
meaningful.

\begin{table}[htb]
\begin{tabular}[]{|l|lll|}
\hline
Function & GA & DPE & EP \\ \hline
F1 & $2.8 \times 10^{-4}$ & $1.1 \times 10^{-11}$ & $3.1 \times 10^{-66}$ \\
F6 & $2.629 \times 10^{-3}$ & $1.479 \times 10^{-9}$ & $5.193 \times 10^{-96}$ \\
\hline
\end{tabular}
\begin{tabular}[]{|l|lll|}
\hline
Function & MEP & MEP+RS & MEP+RS+DM \\ \hline
F1 & $3.3 \times 10^{-71}$ & $3.2 \times 10^{-125}$ & $1.5 \times 10^{-51}$ \\
F6 & $0$ & $0$ & $0$ \\
\hline
\end{tabular}
\caption{Convergence of Various Evolutionary Algorithms
(GA = Genetic Algorithm, DPE = GA with Dynamic Parameter Estimation, EP
= Evolutionary Programming, MEP = Meta-Evolutionary algorithms, RS =
Recorded Step, DM = Directional Mutation)}\label{tab:t1}
\end{table}

\section{Summary and Discussion}

This work clearly shows that meta-evolutionary strategies can be
effective in accelerating the convergence of evolutionary programming
algorithms under certain conditions.

Furthermore, the directional mutation and recorded step do not
significantly degrade this performance on simple problems.  They can
provide highly signficant improvement in convergence speed on problems
which involve long narrow valleys.

The directional search algorithm presented here has a number of clear
advantages over carrying a full covariance matrix with each member of
the population.  These include lower storage requirements, lower
computational overhead, and an intuitively appealing method for doing
meta-mutation.

Although the algorithms describe here perform well on moderately
multi-modal problems such as the Bohachevsky functions, they probably
trade off some of the ability to avoid locally optimal solutions in
return for their ability to explore narrow valleys.  This ability may
need to be recovered for some problems.  One way that this might be
done is to allow only parent/progeny competition.  This would help
avoid the situation where a solution is found which is good enough to
swamp the survivor pool with progeny before more obscure solutions are
found.  Another method for attacking this problem would be introduce
speciation.

In partially or wholly decomposable problems with high dimensionality,
narrow valleys can occur which are aligned with the axes rather than
aligned arbitrarily.  In these cases, it the cost of the directional
mutation algorithm given here might be better spent by keeping a
separate mutation rate for each dimension.  Each of these mutation
rates could be subjected to the self-similar exponential mutation
described in this paper.  Another option would be to keep the
directional mutation, and expand the omni-directional mutation rate to
one mutation rate per parameter.  Whether either of these changes
would actually enhance performance is an open question.

The use of a Cauchy distribution for the directional mutation is also
an intriguing possibility.  As was noted earlier, the use of Cauchy
distributions in problems similar to the long narrow valley examined
here could severely degrade convergence.  This is because of the fact
that multi-variate Cauchy distributions are not rotationally
invariant.  One intriguing option for a hybrid approach is to use a
Cauchy distribution for the directional mutation while retaining a
normal distribution for the omni-directional mutation.  Such a hybrid
would retain the rotationally invariant properties of the algorithm
described here while taking advantage of the desirable aspects of the
use of the Cauchy distribution.

Another interesting avenue for further research is to combine step
recording and directional mutation with more conventional
self-adaptation of mutation rates for each parameter.  This
combination might provide the advantages of recorded step methods when
solving largely separable problems without losing the ability of
directional mutation to deal with narrow non-separable valleys.  The
cost of this hybrid would be the requirement to keep $2n$
meta-parameters with each member of the population instead of the
$n+1$ meta-parameters required by the methods described here.

Overall, step recording and the directional mutation operator
described in this report seem to provide strong advantages for
optimizing certain classes of problems.  The experiments described
here provide an initial indication of how large these advantages can
be.  Further work is needed to characterize the interactions between
these innovative techniques and other methods.

\newpage
\bibliography{refs}

\begin{thebibliography}{Atm91}

\bibitem[Atm91]{fogel}
David B. Fogel; Larry J. Fogel; J.~Wirt Atmar.
\newblock Meta-evolutionary programming.
\newblock In R.R. Chen, editor, {\em Proceedings of the 25th Asilomar
  Conference on Signals, Systems and Computers}, pages 542--545. Pacific Grove,
  CA, 1991.

\bibitem[BR64]{bremermann64}
H.J. Bremermann and M.~Rogson.
\newblock An evolution-type search method on convex sets.
\newblock Technical Report ONR Technical Report, Contracts 222(85) AND
  3656(58), ONR, 1964.

\bibitem[BR65]{bremermann65}
H.J. Bremermann and M.~Rogson.
\newblock Search by evolution.
\newblock In M.~Maxfield, A.~Callahan, and L.~Fogel, editors, {\em Biophysics
  and Cybernetic Systems}, pages 157--167. Spartan Books, Washington D.C.,
  1965.

\bibitem[FK92]{fishman}
G.~S. Fishman and V.~G. Kulkarni.
\newblock Improving monte carlo efficiency by increasing variance.
\newblock {\em Management Science}, 38(10):1432--1444, 1992.

\bibitem[Fog92]{fogel92}
David B. Fogel; Larry J. Fogel; J. Wirt Atmar; Gary~B. Fogel.
\newblock Hierarchic methods of evolutionary programming.
\newblock In {\em Proc. of the First Annual Conference on Evolutionary
  Programming, Evolutionary Programming Society, San Diego, CA}, 1992.

\bibitem[Fog95]{fogel94}
David~B. Fogel.
\newblock {\em Evolutionary Computation: Toward a New Philosophy of Machine
  Intelligence}.
\newblock IEEE Press, New York, NY, 1995.

\bibitem[Fog97]{fogel97}
David~B. Fogel.
\newblock A preliminary investigation into directed mutations in evolutionary
  algorithms.
\newblock In H.-M. Voigt, W~Ebeling, I.~RechenBerg, and H.-P. Schwefel,
  editors, {\em PPSN4}, pages 329--335. Springer Verlag, Berlin, 97.

\bibitem[Sch81]{schwefel}
H.-P. Schwefel.
\newblock {\em Numerical Optimization of Computer Models}.
\newblock John Wiley, Chichester, UK, 1981.

\bibitem[Seb92]{sebald}
A.V. Sebald.
\newblock On exploiting the global information generated by evolutionary
  programs.
\newblock In {\em Proc. of the First Annual Conference on Evolutionary
  Programming, Evolutionary Programming Society, San Diego, CA}, 1992.

\bibitem[SH87]{szu}
H.~H. Szu and R.~L. Hartley.
\newblock Nonconvex optimization by fast simulated annealing.
\newblock {\em Proceedings of the IEEE}, 75:1538--1540, 1987.

\bibitem[Yao91]{yao91}
Xin Yao.
\newblock Simulated annealing with extended neighborhood.
\newblock {\em International Journal of Computer Mathematics}, 40:169--189,
  1991.

\bibitem[Yao95]{yao95}
Xin Yao.
\newblock A new simulated annealing algorithm.
\newblock {\em International Journal of Computer Mathematics}, 56:161--168,
  1995.

\bibitem[YL97]{yao97}
Xin Yao and Yong Liu.
\newblock Fast evolutionary programming.
\newblock In {\em Evolutionary Programming, V}. MIT Press, 1997.

\end{thebibliography}
\end{document}